\documentclass[letterpaper, 10 pt, conference]{ieeeconf}
\IEEEoverridecommandlockouts
\usepackage[hyphens]{url}
\usepackage{afterpage}
\usepackage[noadjust]{cite}
\usepackage[bookmarks=true]{hyperref}
\usepackage{amsmath}
\usepackage{times}
\usepackage{amssymb}
\usepackage{dsfont}
\usepackage{algpseudocode}
\usepackage{algorithm}
\usepackage{subcaption}
\usepackage{graphicx}
\usepackage{float}
\usepackage{booktabs}
\usepackage{ragged2e}
\usepackage{adjustbox}
\usepackage{bm}
\usepackage{multirow}
\usepackage{comment}
\usepackage{overpic}
\usepackage{soul}
\usepackage[table, dvipsnames]{xcolor}
\usepackage{flushend}
\usepackage{balance}

\newtheorem{theorem}{Theorem}

\providecommand{\Opt}{\texttt{(Opt)}}
\providecommand{\optref}{\hyperref[opt]{\Opt{}}}

\title{\LARGE \bf NeHMO: Neural Hamilton-Jacobi Reachability Learning for Decentralized Safe Multi-Arm Motion Planning}

\author{Qingyi Chen$^{1}$, Zachary Kingston$^{1}$ and Ahmed H. Qureshi$^{1}$
\thanks{$^{1}$Qingyi Chen, Zachary Kingston and Ahmed H. Qureshi are with the Department of Computer Science at Purdue University, West Lafayette, IN, USA. {\tt\footnotesize \{chen5221, zkingston, ahqureshi\}@purdue.edu}.}%
}

\begin{document}
\maketitle

\begin{abstract}

Safe multi-arm motion planning is a challenging problem in robotics due to its high dimensionality, coupled configuration space, and complex collision constraints. Centralized planners are capable of coordinating all arms but often face scalability limitations, restricting applicability in real-time settings. On the other hand, decentralized methods are scalable and recent deep learning-based approaches have shown promising results. However, these depend on accurate behavior prediction or coordination protocols and may fail when other arms act unpredictably. To address these challenges, we introduce a neural Hamilton-Jacobi Reachability (HJR) learning-based approach to approximate a safety value function that captures worst-case inter-arm safety constraints. 
We further develop a decentralized trajectory optimization framework that uses the learned HJR representation for real-time planning. The proposed method is scalable and data-efficient, generalizes across multi-manipulator systems, and outperforms state-of-the-art baselines on challenging multi-arm motion planning tasks. 

\end{abstract}
\section{Introduction}

Multi-arm motion planning plays an important role in robotic manipulation systems, including coordinated assembly, shared-workspace manipulation, and warehouse automation. However, ensuring system-wide safety while generating motion plans in real time remains a significant challenge.

One approach is to plan jointly for all robots using a centralized controller \cite{shome2020drrt, solis2021representation, ratliff2009chomp}. Such planners explicitly coordinate all arms and enforce global safety constraints. However, their computational complexity grows rapidly with the number of manipulators \cite{ha2020learningdecentralizedmultiarmmotion}, limiting scalability and real-time applicability. An alternative is decentralized planning, where each arm generates its own plan, improving scalability and computational efficiency at the cost of typically relying on predicting other arms’ behaviors or assuming their compliance with a shared control policy. These assumptions may break down in unstructured or safety-critical settings, particularly when other manipulators behave unpredictably. 

\begin{figure}[!t]
\hskip 2ex
\begin{subfigure}[b]{0.22\textwidth}
    \begin{overpic}[width=\textwidth]{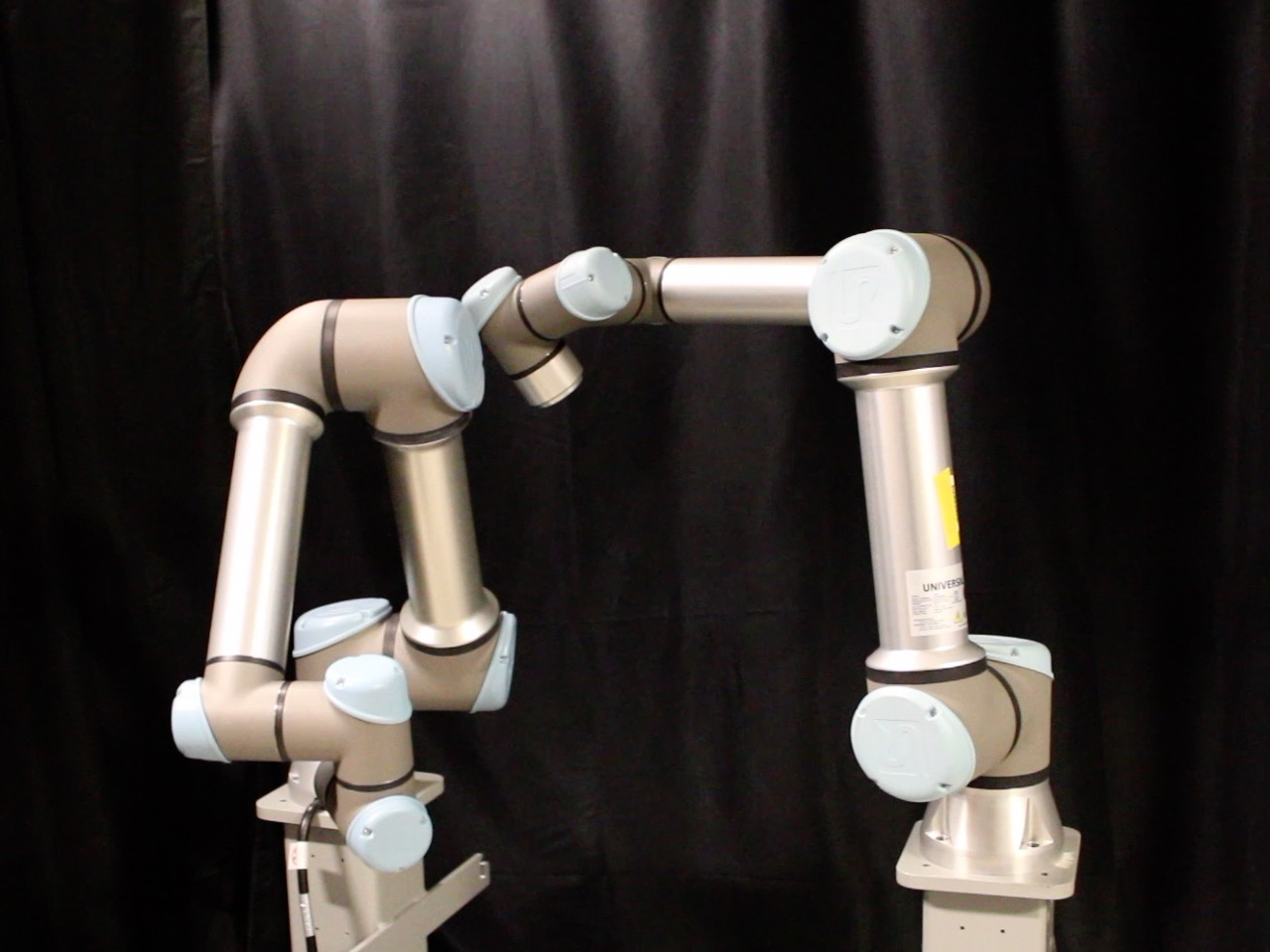}
        \put(90,5){\color{white}\bfseries\small (1)}
    \end{overpic}
\end{subfigure}
\hskip -1ex
\begin{subfigure}[b]{0.22\textwidth}
    \begin{overpic}[width=\textwidth]{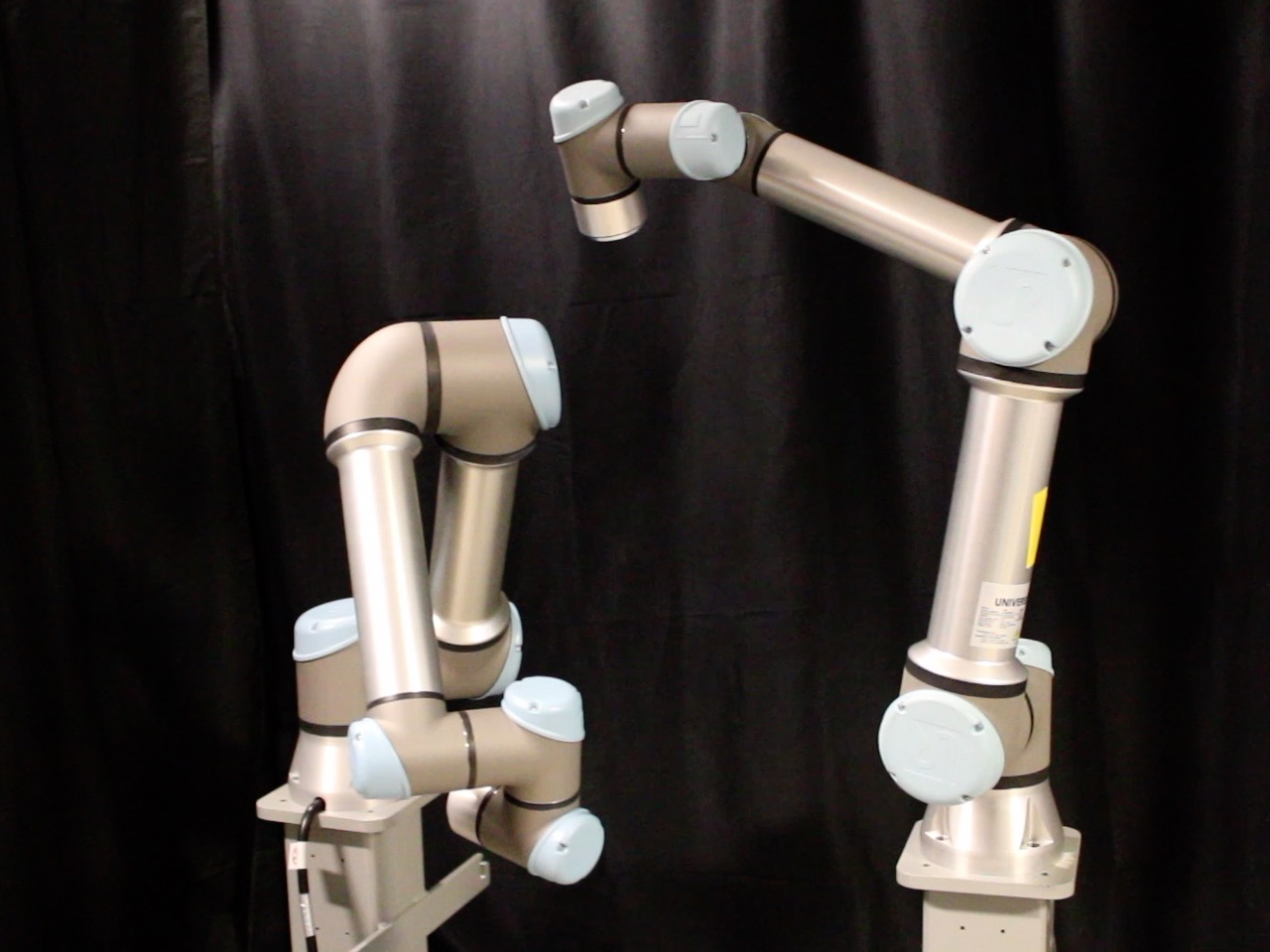}
        \put(88,5){\color{white}\bfseries\small (2)}
    \end{overpic}
\end{subfigure}

\hskip 2ex
\begin{subfigure}[b]{0.22\textwidth}
    \centering
    \begin{overpic}[width=\textwidth]{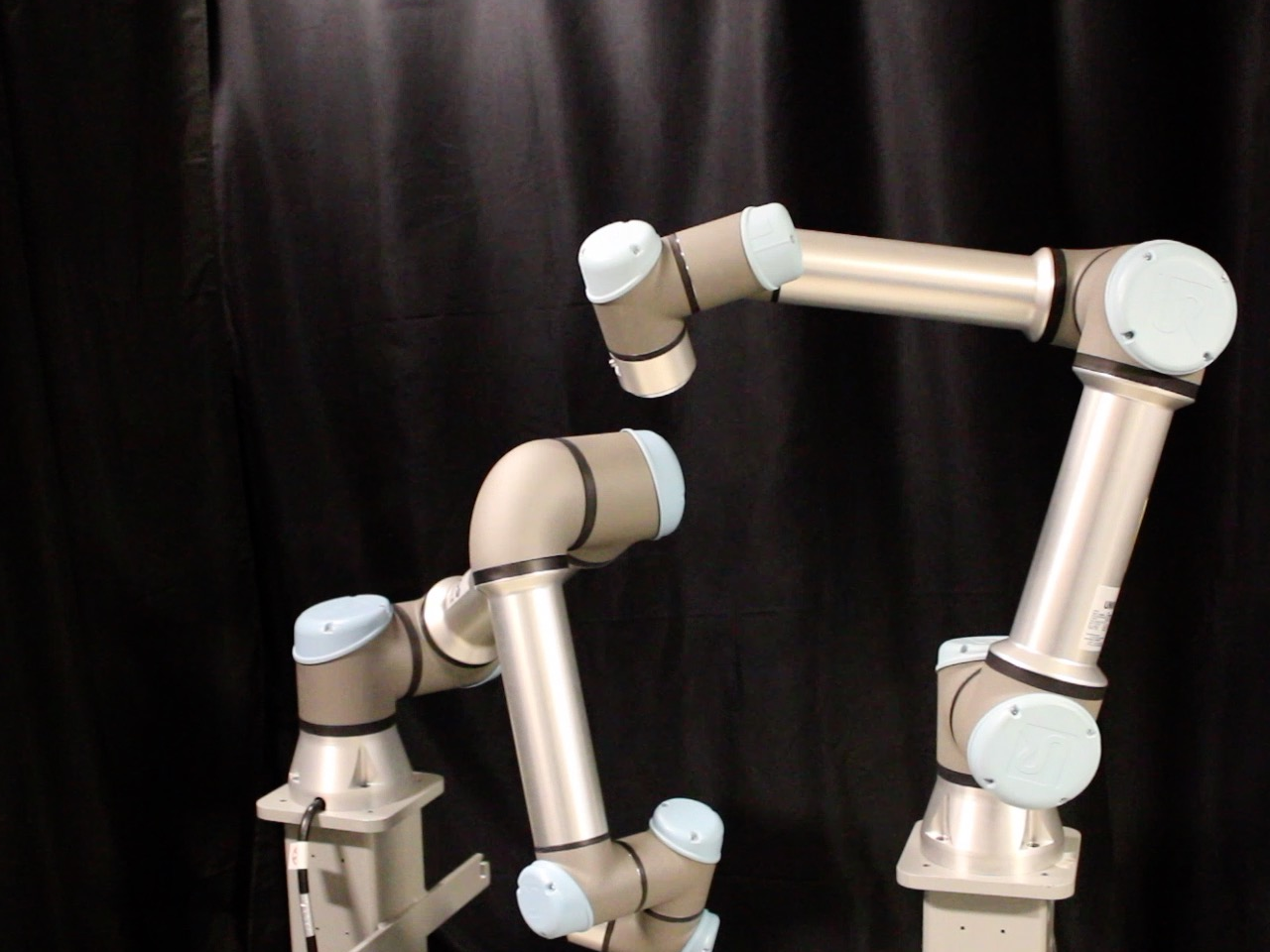}
    \put(90,5){\color{white}\bfseries\small (3)}
    \end{overpic}
\end{subfigure}
\hskip -1ex
\begin{subfigure}[b]{0.22\textwidth}
    \centering
    \begin{overpic}[width=\textwidth]{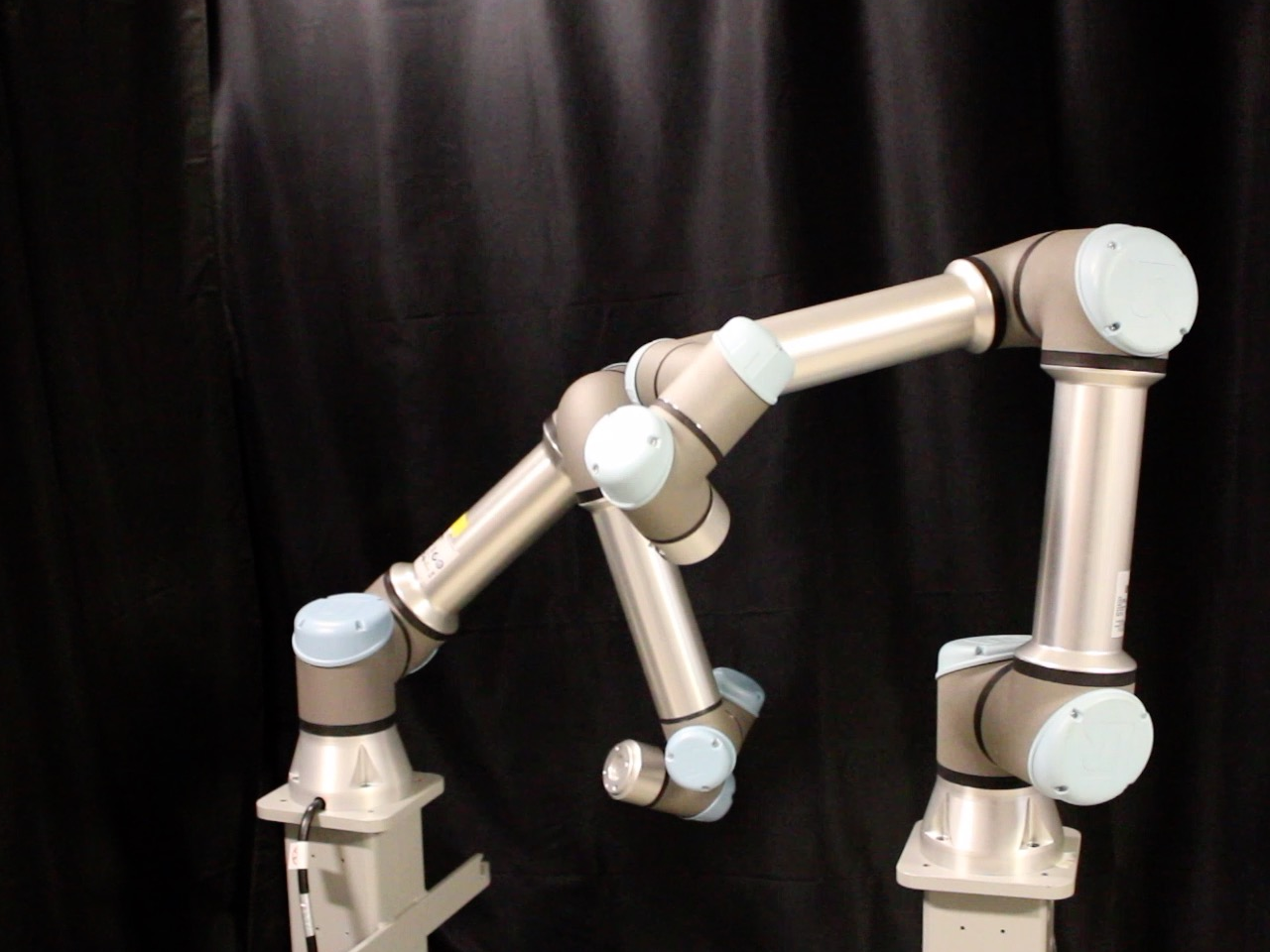}
    \put(88,5){\color{white}\bfseries\small (4)}
    \end{overpic}
\end{subfigure}
\caption{Our method, NeHMO, controlling a 12-dimensional dual-UR5 system. The images show the two manipulators getting close to each other and adjusting their configurations to avoid a potential collision. }
\vspace{-5mm}
\label{fig:intro_fig}
\end{figure}

Hamilton-Jacobi Reachability (HJR) provides a principled framework for modeling safety-critical control and has been increasingly applied to multi-robot systems \cite{recentadvances, multi-vehicle-hjr-mip, infusing-multi}. Based on the Hamilton-Jacobi-Isaacs (HJI) equation, HJR formulates safety as a differential game between control and disturbance, enabling reasoning about worst-case interactions under dynamic constraints. In multi-arm systems, this formulation naturally captures inter-arm collision avoidance under uncertainty or adversarial behavior. However, classical HJR methods suffer from exponential scaling with dimensionality, making direct application to high-dimensional systems infeasible. To address this limitation, Bansal and Tomlin \cite{deepreach} introduced DeepReach, which uses neural networks to approximate HJR solutions through self-supervised training. While this approach removes the need for labeled data, it requires extensive training and does not readily scale to high-dimensional multi-arm systems.

To address these challenges, we present \textbf{Ne}ural \textbf{H}JR-guided \textbf{M}ulti-arm \textbf{M}otion \textbf{O}ptimizer (NeHMO). NeHMO augments DeepReach with physics priors in multi-robot systems, improving the modeling and scalability of neural HJR in higher-dimensional articulated systems. It further integrates the learned HJR representation into a decentralized trajectory optimization framework that efficiently solves complex multi-arm motion planning problems in real time. Our key contributions are summarized as follows:
\begin{itemize}
    \item A training pipeline that augments DeepReach \cite{deepreach} to exploit structural symmetry and solve multi-robot HJR problems with improved data efficiency and scalability.
    \item A decentralized trajectory optimization framework that incorporates the learned HJR representations to actively avoid unsafe multi-arm interactions. 
    \item Empirical validation on multi-arm motion planning tasks up to a 30-DoF system with five manipulators, including heterogeneous manipulators and near-adversarial scenarios.
\end{itemize}

\section{Related Work}
Safe multi-arm motion planning studies the problem of guiding multiple manipulators to their respective goals while avoiding inter-arm and environmental collisions. Traditional centralized motion planning algorithms offer a direct solution by globally coordinating the motions of all manipulators \cite{ratliff2009chomp, shome2020drrt, kuffner2000rrt}.
While dependent on their specific frameworks and algorithms, the runtime of centralized methods often scales inefficiently with the number of agents, making them less practical for real-time applications and large-scale environments \cite{ha2020learningdecentralizedmultiarmmotion}. In contrast, our approach adopts a decentralized structure, allowing each arm to optimize its own trajectory while accounting for safety through learned shared representations rather than explicit global coordination.

Decentralized methods generally offer improved scalability by allowing each agent to plan its own motion \cite{lin2022review}. While many such strategies have been developed for mobile and aerial robots \cite{van2008reciprocal}, extending them to high-dimensional articulated systems with complex collision geometries remains challenging \cite{guo2026efficient}. In particular, robot manipulators introduce configuration-dependent inter-arm collision constraints that are difficult to capture using geometric abstractions designed for simpler platforms. Our method addresses this challenge by learning structured safety representations directly in the manipulators’ configuration spaces, enabling decentralized planning without simplifying the articulated geometry.

Learning-based approaches provide an alternative by amortizing computation through offline training and enabling fast inference at deployment. Notably, Ha et al. \cite{ha2020learningdecentralizedmultiarmmotion} demonstrated successful learning of a multi-arm policy for 6-DoF manipulators. However, training such policies in high-dimensional spaces often requires expert demonstrations or carefully curated datasets. Moreover, although safety filtering mechanisms have been incorporated in some works~\cite{safaoui2024droneRL, shao2021reachability}, purely data-driven policies often lack interpretability and structured safety representations, which limits their robustness in safety-critical settings. Rather than learning an end-to-end control policy, our approach learns structured reachability-based safety representations that can be embedded within an optimization-based planner, preserving interpretability while retaining computational efficiency.

Hamilton-Jacobi Reachability (HJR) offers a principled framework for modeling safety-critical control. Classical numerical solvers \cite{mitchell2007toolbox} have demonstrated effectiveness in computing safety sets for multi-robot systems \cite{bansal2021provablesafemultivehicle}. More recently, learning-based solvers such as \cite{deepreach, hsu2024isaacsiterativesoftadversarial} have shown promise in approximating HJR solutions using neural networks, and HJR has also been incorporated into reinforcement learning frameworks to enhance safety \cite{ganai2024iterative}. These methods, however, are often tied to fixed observation spaces or environment configurations, limiting their adaptability to varying obstacle layouts or different numbers of agents. Therefore, we learn representations that decouple agent interactions and environmental factors, enabling reuse across varying multi-arm configurations and obstacle settings.

Trajectory optimization is widely adopted for its flexibility in incorporating diverse safety constraints, including numerically computed HJR solutions \cite{infusing-hj-planning} and control barrier functions (CBFs) \cite{cheng2020robustcbf}. Recent work has also explored integrating neural safety representations into trajectory optimization for improved scalability \cite{jeong2024parameterizedfastsafetracking, michaux2023reachability, yu2024efficient}. However, constructing such safety constraints that remain effective in decentralized multi-arm settings with complex articulated collision geometry remains an open challenge. Our work builds upon this line of research by learning structured HJR representations tailored to multi-arm systems and embedding them within a decentralized trajectory optimization framework. This combination enables scalable, safety-aware planning for high-dimensional articulated manipulators without relying on explicit coordination assumptions.

\section{Proposed Method}\label{sec:method}
This section presents NeHMO, our proposed framework for multi-arm motion planning, which models and solves HJR for high-dimensional articulated systems and integrates it into decentralized planning.
We investigate the problem of safe multi-arm motion planning, where manipulators are tasked to move toward their goals while avoiding collisions with others. This setup models scenarios such as manipulators performing pick-and-place in a shared workspace. We define the problem under the assumption that each robot can perfectly perceive the states of other robots and know their dynamics, but does not have access to their control policies.

\subsection{Hamilton-Jacobi Reachability Modeling}

Our problem considers a multi-robot setup where we model pair-wise interaction between two robots as a coupled dynamical system. We write $x^i \in \mathcal{X}^i$ as the state of the $i$-th robot where $\mathcal{X}^i$ is its state space. For the dual-robot system, we write $x = (x^1, x^2) \in \mathcal{X}$ where $\mathcal{X} = \mathcal{X}^1 \times \mathcal{X}^2$. 

Specifically, we model one robot as the control input to the coupled system and the other as an adversarial disturbance, capturing potentially conflicting objectives between them. 
Under this formulation, safety analysis takes the form of a Backward Reachable Tube (BRT) problem characterized by the Hamilton-Jacobi-Isaacs (HJI) equation.
BRT problems characterize the set of initial states from which trajectories can reach or avoid a target set \textit{within} a time horizon, making them well suited for modeling continuous-time collision avoidance constraints. 
The need to account for the potentially adversarial behavior of other robots leads to the HJI formulation, which models a zero-sum differential game between a collision-avoiding robot (evader) and an adversarial robot (pursuer). We assume the standard conditions required by the HJI framework, including Lipschitz continuity of the system dynamics, compactness of the control and disturbance sets, and nonanticipative strategies for the control input. We refer interested readers to \cite{fisac2014reachavoidproblemstimevaryingdynamics} for a comprehensive discussion of reach-avoid differential games.

Formally, we consider control-affine systems that evolve under the dynamics
\begin{equation}
    \dot{x}(s) = g(x(s),u(s),d(s)),\quad s\in[t,T]
\end{equation}
where the control input $u(\cdot)$ and the disturbance $d(\cdot)$ are measurable functions taking values in compact sets $\mathcal{U}$ and $\mathcal{D}$, respectively. We write $\xi_{u, d, t, x}(s)$ as the trajectory of the system starting from the initial state $x$ and time $t$ under control $u(\cdot)$ and $d(\cdot)$.
To model the safety (collision-avoidance) requirement in multi-robot systems, we consider a collision set $\mathcal{L} = \{x\hspace{0.1cm} |\hspace{0.1cm} l(x) \le 0\}$, where $l(x)$ is typically a signed distance function that differentiates collision and non-collision states by its sign. 
Our goal is to compute a collision-avoiding value function $V(t, x)$ and its corresponding avoid set $\mathcal{V}(t)$, characterized via the Hamilton-Jacobi-Isaacs (HJI) backward reachable tube (BRT) formulation, defined as
\begin{align}
    \mathcal{V}(t) &= \{x\ |\ \exists u(\cdot), \forall d(\cdot), \forall s \in [t, T], \xi_{u, d, t, x}(s) \not \in \mathcal{L}\} \\
      &= \{x\ | \ V(t, x) > 0\}.
\end{align}
Following this formulation, the zero level set $\{x\mid V(t,x) = 0\}$ characterizes the boundary of a safe set that separates states from which collision can be avoided under worst-case disturbances from those for which collision is inevitable.

The value function $V(t, x)$ can be solved as the viscosity solution of the HJI partial differential equation (PDE) \cite{recentadvances,mitchell2002application,mitchell2005time} 
\begin{align}\label{hjipde}
    \frac{\partial V(t, x)}{\partial t} &+ \min\{H(t, x, \nabla_xV(t,x)), 0\} = 0 \nonumber\\
    &V(T, x) = l(x).
\end{align}
Here $H$ is the Hamiltonian of the system, defined as
\begin{align}
     H(t, x, \nabla_xV(t,x)) &= \max_{u}\min_{d}\langle \nabla_xV(t,x),{g(x,u,d)}\rangle
\end{align}
with $\langle \cdot, \cdot\rangle$ denoting a vector inner product and $\nabla_xV(t,x)$ being the gradient of $V(t, x)$ with respect to the state variable $x$, also known as the costate \cite{recentadvances}. 
For control-affine systems, the Isaacs condition holds, guaranteeing existence of a unique value function. Notably, the value function also allows us to derive the optimal control for the evader:
\begin{equation}\label{hjiaction}
    u^*(t, x) = \arg\max_{u}\min_{d}\langle \nabla_xV(t,x),{g(x,u,d)}\rangle.
\end{equation}
The corresponding optimal control of the pursuer can be acquired similarly.

\subsection{Physics-Informed HJR Learning}
A major challenge in HJR is to solve the PDE presented in~\eqref{hjipde}. DeepReach \cite{deepreach} is a leading approach to approximate this solution using neural networks, addressing the exponential scaling of traditional grid-based methods. DeepReach learns the value function in a self-supervised manner by pre-training the neural network to learn the boundary condition at the terminal time and then propagating backward in time to optimize the network and its gradient to satisfy~\eqref{hjipde}. We refer interested readers to \cite{deepreach} for further details on DeepReach.  

While the self-supervised training pipeline in DeepReach removes the need for ground-truth solutions for training, it typically requires extensive sampling to cover the state space and converge to a valid HJR solution that both satisfies the boundary condition and adheres to the governing differential equation. In high-dimensional multi-arm settings, this requirement can lead to substantial training effort. Therefore, rather than treating the problem as structure-agnostic, we augment DeepReach with problem structure inherent to articulated multi-arm systems, motivated by two observations. First, safety boundary conditions often exhibit complex geometric structure when expressed in configuration space due to intricate inter-arm collision geometries. Directly learning the full value function in such spaces can therefore be challenging. Second, multi-robot systems often possess physical and geometric regularities that can be incorporated as prior knowledge during training. Building on these observations, we introduce two priors to improve data efficiency and solution quality during HJR modeling:
(i) exploiting symmetry properties arising from equivalent multi-robot states, and
(ii) explicitly incorporating the boundary condition and learning only the time-dependent residual of the value function.
The following subsections detail how we adapt DeepReach to integrate these two physics-informed priors.

\subsubsection{Symmetry in HJR} 

\begin{figure}[t]

    \begin{subfigure}[t]{0.22\textwidth}
    \includegraphics[trim={0cm 1.2cm 0cm 1.2cm},clip, width=1\linewidth]{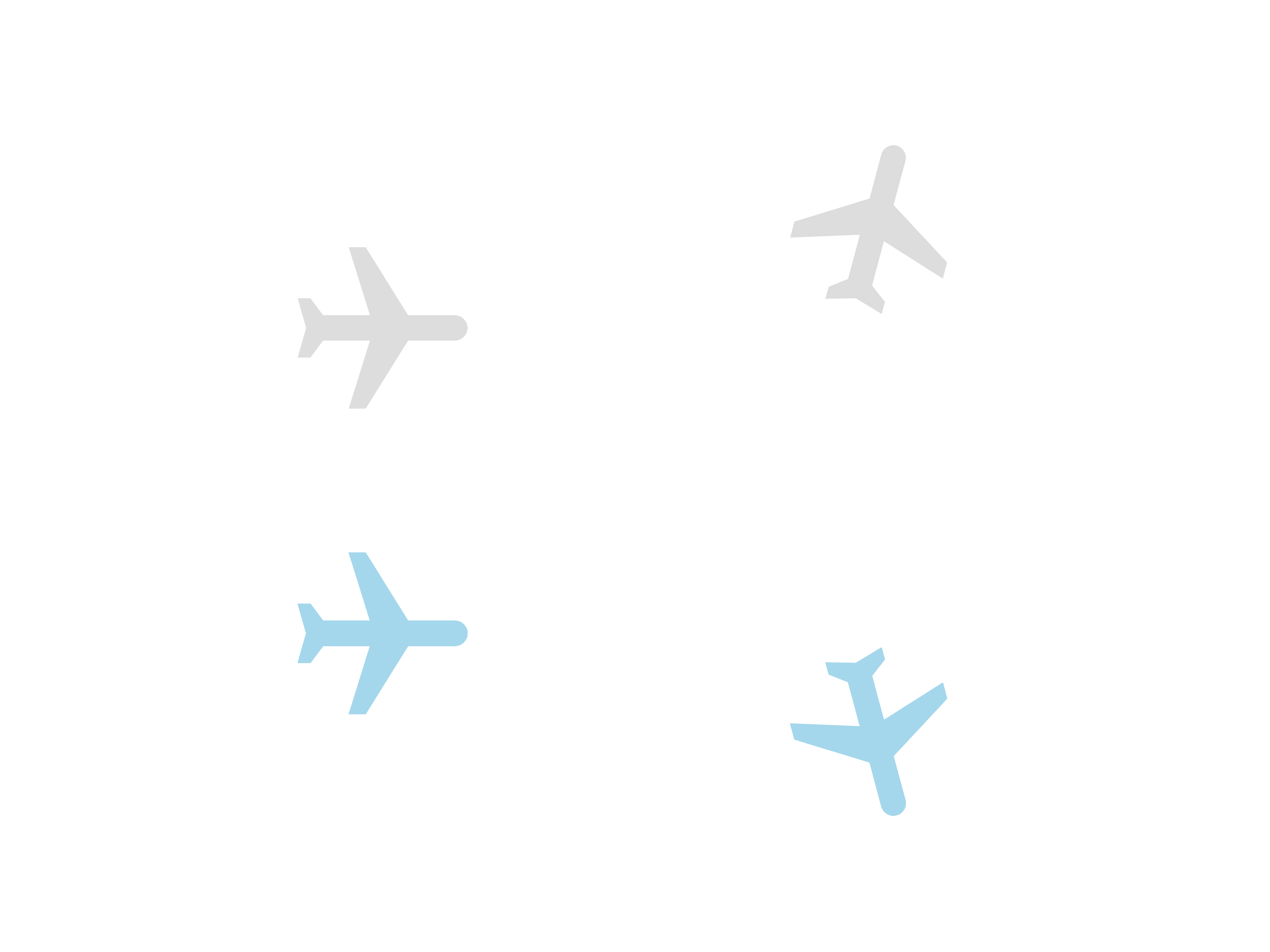}
    \caption{Air3D.}
    \end{subfigure}
    \hskip 1ex
    \begin{subfigure}[t]{0.18\textwidth}
    \includegraphics[trim={2cm 2cm 2cm 2cm},clip, width=1\linewidth]{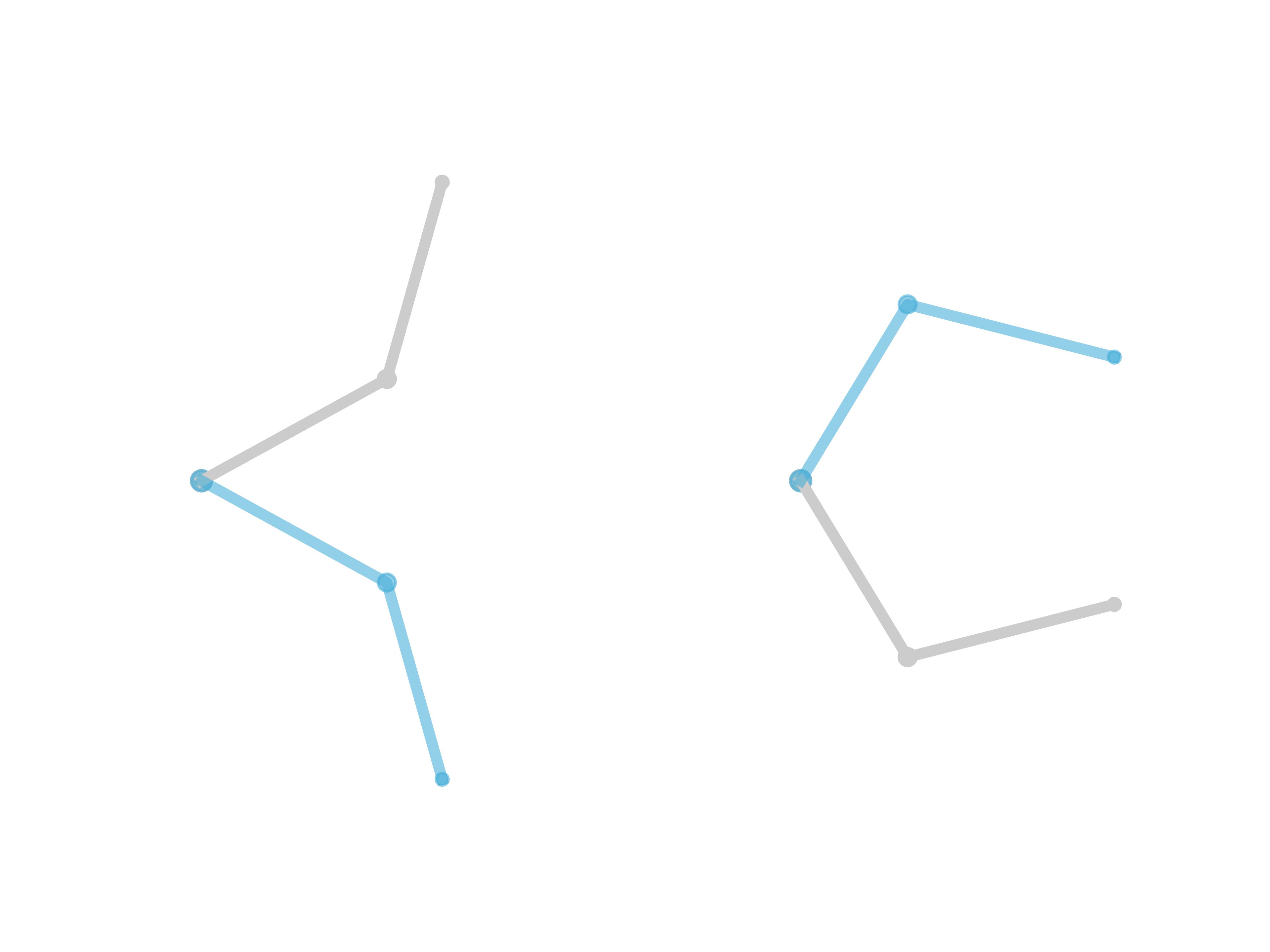}
    \caption{Simple Arms.}
    \end{subfigure}
    \vskip 1ex
    \begin{subfigure}[t]{0.22\textwidth}
    \includegraphics[width=1\linewidth]{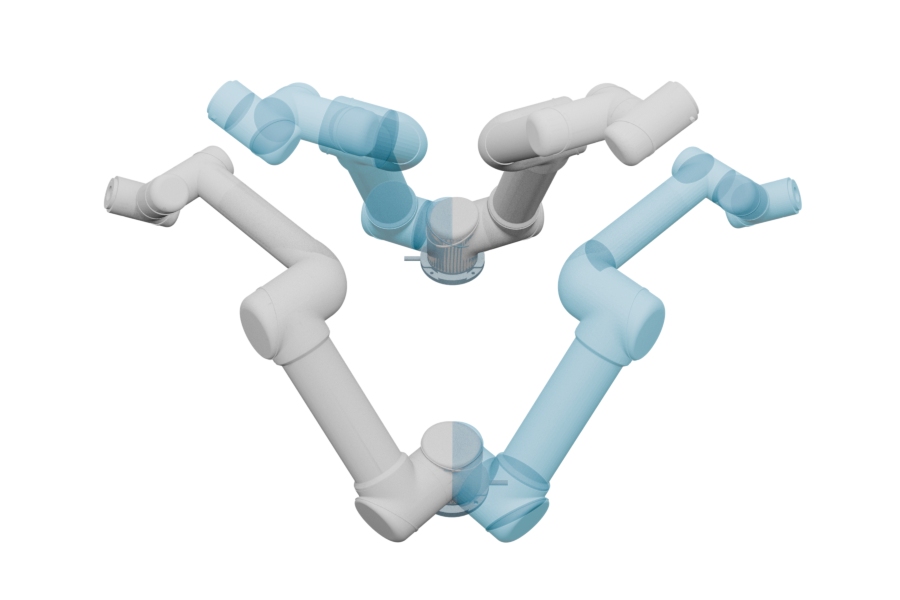}
    \caption{UR5 Manipulators.}
    \end{subfigure}
    \begin{subfigure}[t]{0.22\textwidth}
    \includegraphics[width=1\linewidth]{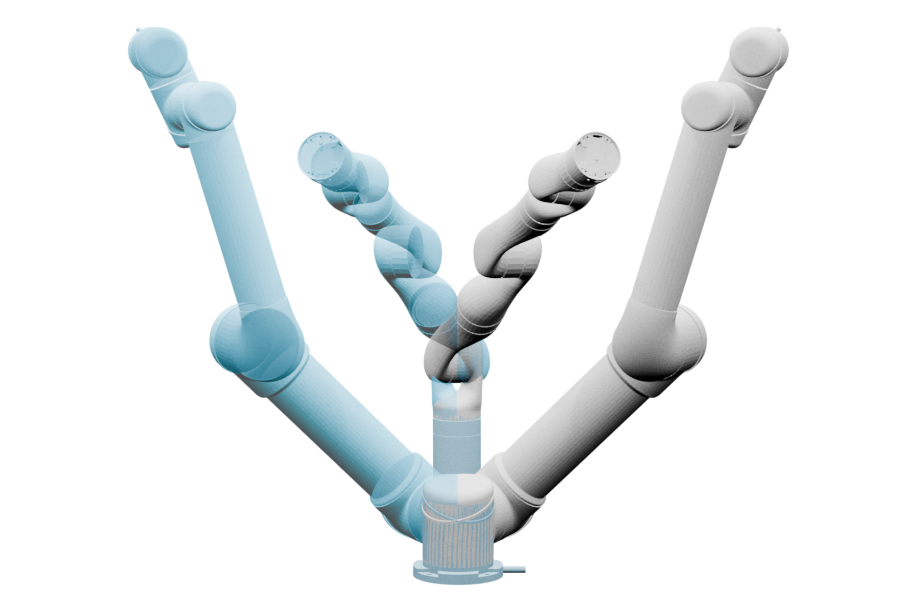}
    \caption{UR5-Kinova Manipulators.}
    \end{subfigure}

    \caption{System setups and their symmetries. The primal state is depicted in gray; its corresponding symmetric state is drawn in blue (transparent).}
    \label{fig:symmetry_figure}
\end{figure}

Multi-robot systems often exhibit a symmetry property that arises from robot geometries and equivalent states, as intuitively exemplified in Figure \ref{fig:symmetry_figure} using several multi-robot systems. Details on these systems and their dynamics can be found in Appendix~\ref{sec:system-setups}. In HJR, such symmetry induces invariance in the value function.

\begin{theorem}
Let $f:\mathcal{X}\rightarrow\mathcal{X}$ be a differentiable bijection on the state space with invertible Jacobian $Df(x)$ such that
\begin{equation}
    l(x) = l(f(x)),\ H(t, x, p) = H(t, f(x), Df(x)^{-\!\top
}p)
\end{equation}
 $\forall\ t\in[0,T],\ x\in \mathcal{X},\ p\in \mathbb{R}^{n_d}$ where $n_d = \dim(\mathcal{X})$. If the associated HJI PDE admits a unique viscosity solution, then the value function satisfies
\begin{equation}
    V(t, x) = V(t, f(x))\ \forall\ t\in[0,T], x \in \mathcal{X}.
\end{equation}
\end{theorem}
This result follows from the uniqueness of viscosity solutions to the HJI equation, as $V(t,x)$ and $V(t,f(x))$ satisfy the same PDE and boundary condition. Intuitively, when $f$ represents a symmetry in the dynamics and the boundary condition, the value function inherits that symmetry.

This symmetry property allows training to be restricted to a representative subset of the state space. During training, we partition the state space $\mathcal{X}$ into two subsets: $\mathcal{X}_{\text{train}}$ and $\mathcal{X}_{\text{sym}}$, such that there exists a differentiable bijective state mapping function $f(x): \mathcal{X}_{\text{train}} \rightarrow \mathcal{X}_{\text{sym}}$ and $\mathcal{X}_{\text{train}} \cup \mathcal{X}_{\text{sym}} = \mathcal{X}$. Training samples are then drawn only from $\mathcal{X}_{\text{train}}$. During inference, the following equation is used to ensure the value function is correctly evaluated across the entire state space:
\begin{equation} \label{eq:sym}
V(t, x) = \begin{cases}
    V(t, x) &\text{ if } x \in \mathcal{X}_{\text{train}}\\
    V(t, f^{-1}(x)) &\text{ if } x \in \mathcal{X}_{\text{sym}}\\
\end{cases}
\end{equation}
For $x\in \mathcal{X}_{\text{sym}}$, the gradient $\nabla_x V(t,x)$  can be computed using the chain rule. This reduces sampling requirements while preserving correctness of the value function.

\subsubsection{Boundary Condition Residual Learning}
Rather than learning the full value function $V(t, x)$, we target the learning objective to be the time-dependent residual 
\begin{equation} \label{eq:bc}
V_{\text{res}}(t, x) = V(t, x) - l(x)
\end{equation}
where $l(x)$ represents the boundary condition. This decomposition separates geometric boundary structure from temporal dynamics, reducing the function complexity that must be represented by the neural network. While this formulation was discussed in \cite{singh2024exactimpositionsafetyboundary} as a baseline and yielded limited empirical improvement, we observe that for articulated multi-arm systems where collision sets induce complex boundaries in configuration space, this residual parameterization substantially improves training convergence and solution quality. 

\subsection{Trajectory Optimization}
Although the learned value function $V_\theta(t, x)$ captures inter-arm safety interactions, collision avoidance alone does not accomplish the full multi-arm motion planning objective. We therefore incorporate $V_\theta(t, x)$ as a safety constraint within a trajectory optimization layer that steers each arm toward its goal while promoting safe behavior. To account for uncertain behaviors of other manipulators, we also use ~\eqref{hjiaction} together with $V_\theta(t, x)$ to infer the worst-case actions of other robots, enabling each arm to generate defensive motion plans without assuming access to their control policies. This formulation allows the planner to operate under uncertain or unknown arm behaviors and improves robustness in dynamic multi-arm environments.

At each planning step, the arm’s motion plan is computed by solving a receding-horizon trajectory optimization problem.
It plans over a horizon of $t_{\text{plan}}$ while aiming to maintain safety for a duration of at least $t_{\text{safe}}$ after executing the plan. Pairwise inter-arm safety constraints are evaluated using the learned HJR value function. The resulting optimization problem \optref is formulated as follows:
\begin{align} \label{opt}
    & \min_{u\in \mathcal{U}} \texttt{cost(} x^1(t_{\text{plan}}), u, x^1_{\text{goal}}\texttt{)} \\  
    \textbf{s.t.} & \hspace{0.15cm} x^{1}(t+\Delta t) = g_d(x^1(t), u)\\
    & \hspace{0.15cm} x^{i}(t+\Delta t) = g_d(x^i(t), d^{i})\\
    & \hspace{0.15cm} \tau = T - t_{\text{plan}} \\
    & \hspace{0.15cm} d^{i} = \arg\min_{d\in \mathcal{D}}\max_{u\in \mathcal{U}}\langle \nabla_x V_\theta(\tau,x^1(t), x^i(t)), g(x(t),u,d) \rangle \label{eq:disturbance} \\
    & \hspace{0.15cm} V_\theta(T - t_{\text{safe}}, x^{1}(t_{\text{plan}}), x^{i}(t_{\text{plan}})) > \epsilon  \\
    & \phantom{---------------} \text{ for } i=2,\cdots,m \nonumber
\end{align}
where $m$ is the number of robots in the environment, $\epsilon > 0$ is a safety margin, and $g_d(x^i(t), u)$ denotes the discrete-time dynamics model. We assume that the control input stays constant over the planning horizon $t_{\text{plan}}$. The hyperparameters $\epsilon$, $t_{\text{plan}}$, and $t_{\text{safe}}$ are user-defined. In our implementation, we use a quadratic terminal cost 
\begin{equation}
    \texttt{cost(} x^1(t), u, x^1_{\text{goal}}\texttt{)} = \|x^1(t_{\text{plan}}) - x^1_{\text{goal}} \|_2^2
\end{equation}
and choose $t_{\text{plan}}=0.1$, $t_{\text{safe}}=0.3$, and $\epsilon=0.05$. We do not wrap the continuous joints in the cost function to encourage the manipulators to take a longer route and increase the interaction between manipulators. 
Static obstacles can be incorporated by adding additional constraints within the same formulation. Since the system is velocity-controlled and the admissible inputs are box-constrained, the worst-case disturbance in~\eqref{eq:disturbance} admits a closed-form solution, enabling efficient optimization.

\subsection{Decentralized Safe Multi-agent Motion Planning}
Algorithm \ref{alg:online_planning} outlines the planning procedure. At each planning step, the method acquires the current configurations of all agents in the environment to set up the optimization problem \optref. The resulting solution provides the control input for the next planning horizon. In the event that the optimization problem cannot be solved within the allocated time, we apply a fail-safe strategy that takes the most conservative behavior (i.e.,~\eqref{hjiaction}) to avoid collision.

The robot continuously executes its current motion plan while simultaneously computing the motion plan for the next horizon. This enables it to adapt its motion plan in real-time as the environment and other robots' states evolve.
\begin{algorithm}[b]
\begin{algorithmic}[1]
    \State \textbf{Require:} $0 < t_{\text{plan}}, t_{\text{safe}} < T$, $V_\theta$, $f_d$
    \State \textbf{Loop:} \Comment{\textit{Line 9 executes in parallel with Lines 3-8}}
    \State \hspace{0.2cm} $x^1(t) \gets$ \verb|UpdateSelfState()|
    \State \hspace{0.2cm} $\{x^i(t)\}^m_{i=2} \gets$ \verb|PerceiveOtherAgents()|
    \State \hspace{0.2cm} {\bf Try} 
    \State \hspace{0.4cm} $u^* \gets$ \verb|Opt(|$x^1(t), \{x^i(t)\}^m_{i=2}, x^1_{\text{goal}}, t_{\text{plan}}, t_{\text{safe}},\epsilon$\verb|)|
    \State \hspace{0.2cm} \textbf{Catch} \verb|OptInfeasible| or \verb|OptTimeout|
    \State \hspace{0.4cm} $u^*\gets$ \verb|FailSafeControl()|
    \State \hspace{0.2cm} {\bf Execute} $u^*$
    \State {\bf End}
\end{algorithmic}
\caption{\texttt{plan(}$x^1(t), \{x^i(t)\}^m_{i=2}, x^1_{\text{goal}}, t_{\text{plan}}, t_{\text{safe}}, \epsilon$\texttt{)}}
\label{alg:online_planning}
\end{algorithm}

\section{Experiments and Demonstrations} \label{sec:exp}
This section presents three sets of analyses to evaluate the proposed NeHMO framework. First, we study the effect of symmetry exploitation and explicit boundary inclusion on learning HJR solutions with neural networks. This evaluation is conducted on the Air3D system and a 2-link planar arm system, where numerical HJR solutions are available for comparison. Higher-dimensional manipulators (e.g., UR5) are excluded from this analysis due to the absence of tractable ground-truth HJR solvers in high-dimensional spaces. These systems' detailed setups and dynamics are provided in Appendix~\ref{sec:system-setups}. Second, we evaluate NeHMO on multi-arm motion planning tasks. We compare against relevant baselines on a 12-DoF dual-UR5 platform and further examine scalability on 18-DoF (three UR5 arms) and 30-DoF (five UR5 arms) systems. Third, we stress-test NeHMO in challenging interaction scenarios. We consider a heterogeneous dual-arm setup consisting of a UR5 manipulator and a Kinova Gen3 manipulator. In addition to difficult start-goal configurations that induce tight interactions, we evaluate safety performance when the Kinova arm executes a reckless policy that ignores other robots in the environment. This setup approximates a near-adversarial setting and tests the robustness of our decentralized planner.
    
We use the following metrics for the evaluation in motion planning experiments: 
\begin{itemize}
    \item Success Rate (\textbf{SR\%}): Percentage of trials in which all manipulators reach their goals without collision.
    \item Collision Rate (\textbf{CR\%}): Percentage of trials where any inter-arm collision occurs at any time during execution.
    \item Planning Time (\textbf{Time}): Mean computation time for each robot to generate a plan (i.e., to execute Algorithm \ref{alg:online_planning}).
    \item Mean Path Length (\textbf{MPL}):  Mean trajectory length over successful trials. Since our worst-case interaction modeling may induce conservative behaviors, this metric quantifies the efficiency trade-off and is compared against expert-generated trajectories.
\end{itemize}
Finally, all experiments are conducted on a desktop with Intel(R) Xeon(R) w5-2455X CPUs and an NVIDIA GeForce RTX 4070 GPU. The optimization problem \optref is solved with \texttt{IPOPT} \cite{wachter2006implementation}. The multi-arm environment is adapted from the work in \cite{Michaux-SPARROWS-RSS-24}.

\subsection{HJR Learning Validation}
This section evaluates the proposed physics-informed training pipeline for learning neural approximations of HJR solutions. We compare three variants: (i) \textbf{DeepReach}, the original training pipeline, (ii) DeepReach with explicit boundary conditions (\textbf{bc}), which only learns $V_{\text{res}}(t,x)$ defined in~\eqref{eq:bc}, and (iii) DeepReach with explicit boundary conditions and symmetry exploitation (\textbf{bc+sym}), our full method, which combines the residual formulation with symmetry reduction and trains only on $x\in \mathcal{X}_{\text{train}}$. Because symmetric states are not used during training, \textbf{bc+sym} uses only half the training data compared to \textbf{DeepReach} and \textbf{bc}.

Table \ref{tab:validation-loss} reports the mean $l1$-error on a validation set, averaged over three runs with different random seeds. Each run trains \textbf{DeepReach} and \textbf{bc} with 80 million samples and trains \textbf{bc+sym} with 40 million samples; all other hyperparameters are identical. The validation sets are obtained from numerical solvers using \texttt{hj\_reachability} \cite{Schmerling_reachability} and \texttt{pytorch\_kinematics} \cite{Zhong_PyTorch_Kinematics_2024}. For higher-dimensional systems (e.g., UR5), numerical ground-truth solutions are intractable; their evaluation is therefore conducted indirectly through motion planning experiments in Section~\ref{sec:ur5-mp}. 

Including the explicit boundary term yields modest improvement on the Air3D system, where the boundary is relatively simple, but significantly improves solution quality for the 2-link arm system, whose articulated configuration space induces more intricate collision boundaries. 

To further assess symmetry exploitation, we compare \textbf{bc} and \textbf{bc+sym} under varying data usages. In each training iteration, a training batch $\mathcal{X}_{\text{batch}}$ is split into $\mathcal{X}_{\text{batch}}\cap \mathcal{X}_{\text{train}}$ and $\mathcal{X}_{\text{batch}}\cap \mathcal{X}_{\text{sym}}$, where the latter corresponds to states that can be inferred via symmetry and is therefore discarded during training. For example, in Air3D where $V(t, x_1, x_2, x_3) = V(t, x_1, -x_2, -x_3)$, we only train on the subset where $x_3 \ge 0$. Consequently, \textbf{bc+sym} is trained with half the data of \textbf{bc}. Still, as shown in Table~\ref{tab:validation-loss}, \textbf{bc+sym} consistently matches the performance of \textbf{bc} while using fewer samples, demonstrating improved data efficiency.

{
\renewcommand{\arraystretch}{0.85}
\begin{table}[t]
    \centering
    \begin{adjustbox}{width=\columnwidth,center}
    \begin{tabular}{ccccccc}
    \toprule
    \multirow{2}{*}{Systems} & \multirow{2}{*}{Methods} & \multicolumn{5}{c}{Data Usage} \\
    & & 10M & 20M & 40M & 60M & 80M \\\midrule
    \multirow{3}{*}{Air3D} & DeepReach & 0.187 & 0.124 & 0.035 & 0.016 & 0.015 \\
     & bc & 0.074 & 0.051 & 0.027 & 0.017 & 0.014 \\
     & bc+sym &  \textbf{0.054} & \textbf{0.028} & \textbf{0.014} & - & - \\\midrule
    \multirow{3}{*}{Simple Arm } & DeepReach & 0.125 & 0.046 & 0.024 & 0.017 & 0.016 \\
    & bc & \textbf{0.052} & 0.051 & 0.040 & 0.011 & 0.010 \\
    & bc+sym & \textbf{0.052} & \textbf{0.044} & \textbf{0.010} & - & - \\
    
    \bottomrule
    \end{tabular}
    \end{adjustbox}
    \caption{Validation error of each variant at different data usages. }
    \label{tab:validation-loss}
\end{table}
}

\subsection{Multi-UR5 Motion Planning}
\label{sec:ur5-mp}
This section evaluates our planning framework on multi-UR5 motion planning tasks.

\subsubsection{Dual-UR5 Motion Planning} \label{sec:dual-ur5}
We first simulate a reaching task in which two UR5 manipulators aim to reach their respective goal poses without colliding. We generate 100 scenarios by randomly sampling start and goal configurations and ensure that all sampled states are initially collision-free. The UR5 robots are placed 0.6 meters apart to create a shared workspace region that induces frequent interaction and nontrivial collision-avoidance behavior.

\subsubsection{Comparisons and Baselines}
We compare \textbf{NeHMO} on the dual-UR5 setup against the following baselines: 
\begin{itemize}
    \item \textbf{CHOMP} \cite{ratliff2009chomp}, using the implementation from \cite{coleman2014reducingbarrierentrycomplex}, as a centralized trajectory optimization method that produces expert-quality trajectories;
    \item \textbf{Ha et al.}'s  \cite{ha2020learningdecentralizedmultiarmmotion}, a decentralized multi-arm planner trained via reinforcement learning;
    \item \textbf{\Opt +DeepReach}, which replaces our physics-informed HJR model with the original DeepReach solution within the same optimization framework;
    \item \textbf{Naive Planner}, which  drives each arm toward its goal without accounting for other manipulators.
\end{itemize}
We additionally report an ablation, \textbf{NeHMO (w/o Sym)}, which removes symmetry reduction while retaining the boundary residual formulation. This evaluates the impact of symmetry exploitation in the dual-UR5 setting.


\subsubsection{Experiment Results}
Table~\ref{tab:arm_planning_exp} summarizes the results. NeHMO achieves the highest success rate and lowest collision rate, with CHOMP performing competitively due to its centralized global coordination. However, its centralized formulation results in longer computation time, limiting applicability in dynamic environments. 
The conservative formulation used in NeHMO yields approximately 27\% longer paths than CHOMP, as it is designed to account for worst-case behaviors of other agents and prioritize safety under unpredictable interactions. For similar reasons, NeHMO may become stuck when the goals of the two manipulators are close, resulting in trials that are neither successful nor colliding. Additionally, for NeHMO and other optimization-based planners such as CHOMP, continuous joints are not wrapped in the cost function, which leads to longer path lengths. The RL-based method of Ha et al. is not affected by this representation choice. 
Despite solving a nonlinear trajectory optimization problem at each planning step, NeHMO operates in real time at approximately 25Hz, demonstrating practical efficiency for high-dimensional dual-arm systems. 

\begin{figure}[t]
\centering
\hskip -1ex
\begin{subfigure}[t]{0.18\textwidth}
\includegraphics[trim={2cm 0cm 2cm 0cm},clip,width=1\textwidth]{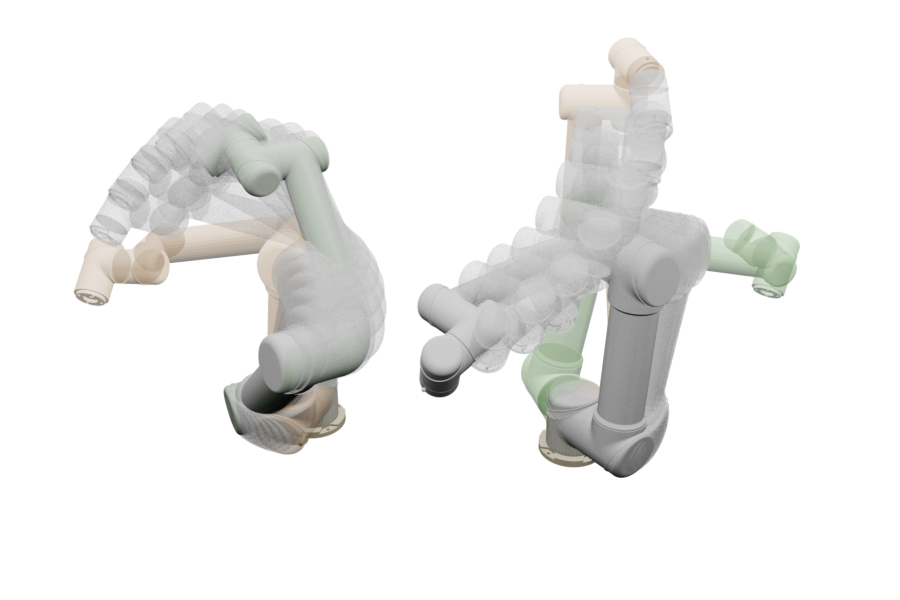}
\vskip -4ex
\caption{}
\end{subfigure}
\hskip -4ex
\begin{subfigure}[t]{0.18\textwidth}
\includegraphics[trim={2cm 0cm 2cm 0cm},clip,width=1\textwidth]{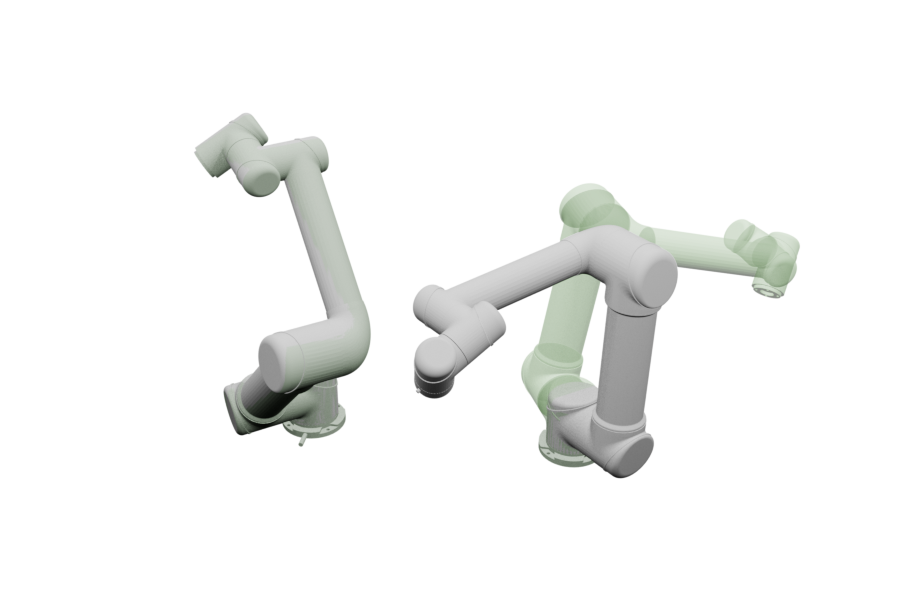}
\vskip -4ex
\caption{}
\end{subfigure}
\hskip -4ex
\begin{subfigure}[t]{0.18\textwidth}
\includegraphics[trim={2cm 0cm 2cm 0cm},clip,width=1\textwidth]{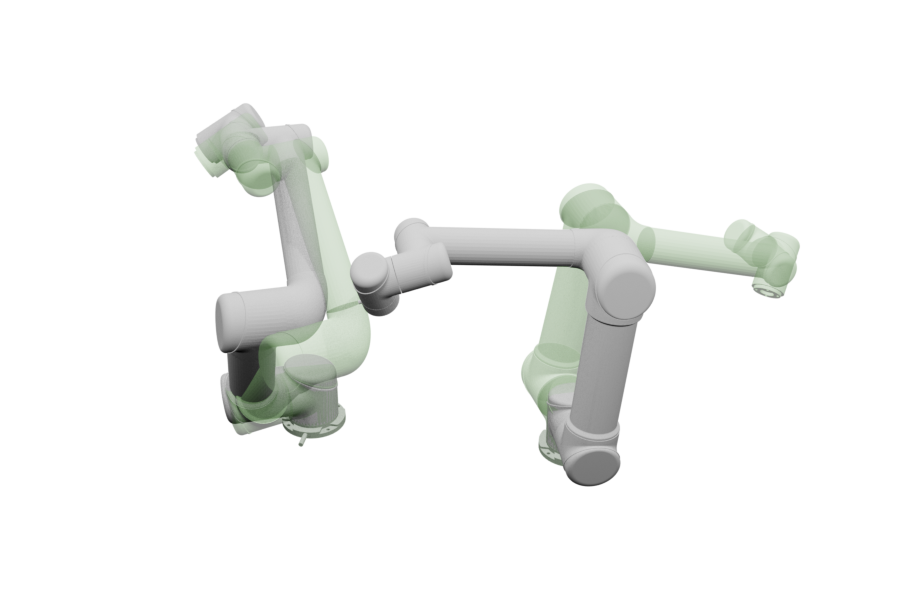}
\vskip -4ex
\caption{}
\end{subfigure}

\vskip -3ex
\hskip -1ex
\begin{subfigure}[t]{0.18\textwidth}
\includegraphics[trim={2cm 0cm 2cm 0cm},clip, width=1\textwidth]{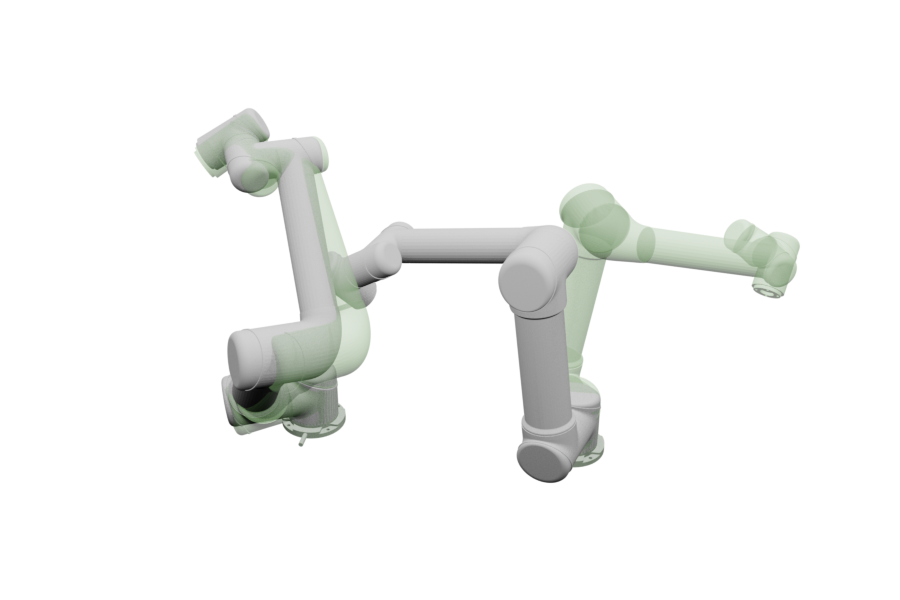}
\vskip -4ex
\caption{}
\end{subfigure}
\hskip -4ex
\begin{subfigure}[t]{0.18\textwidth}
\includegraphics[trim={2cm 0cm 2cm 0cm},clip, width=1\textwidth]{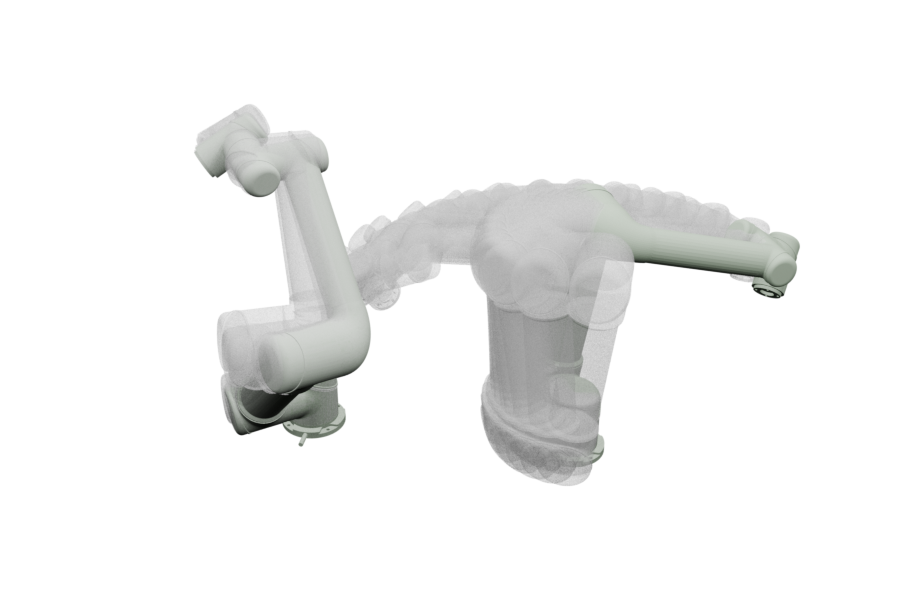}
\vskip -4ex
\caption{}
\end{subfigure}
\hskip -4ex
\begin{subfigure}[t]{0.18\textwidth}
\includegraphics[trim={2cm 0cm 2cm 0cm},clip, width=1\textwidth]{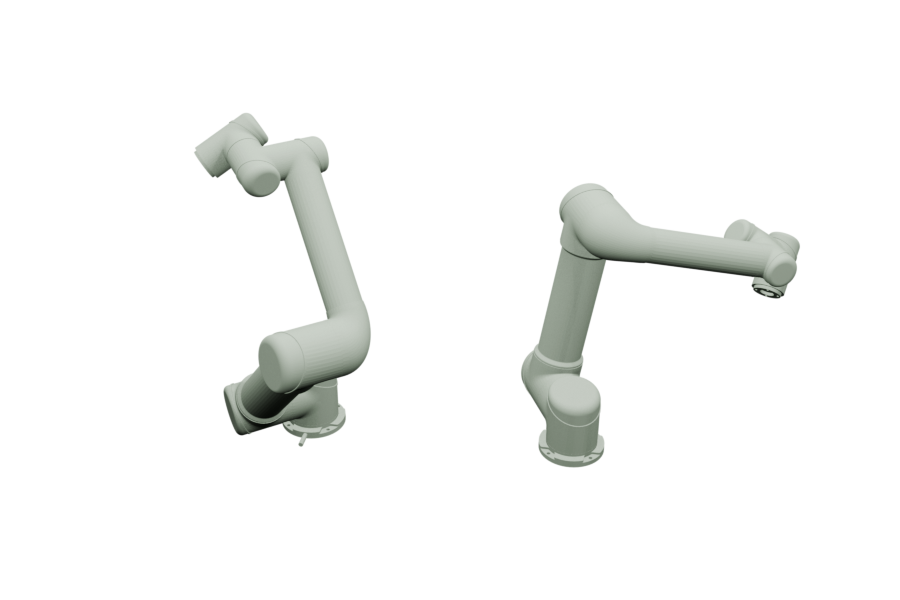}
\vskip -4ex
\caption{}
\end{subfigure}
\vskip -1ex
\caption{An example of NeHMO planning for the dual-UR5 system in simulation.  Current states are gray; intermediate trajectories are transparent; initial configurations are orange; goal configurations are green. The two manipulators start and get close to each other (a-b), adjust their configurations to avoid a potential collision (c-d), and then safely reach their goals (e-f).}
\vskip -2ex
\label{fig:planning_fig2}
\end{figure}

\begin{table}[b]
    \centering
\begin{adjustbox}{width=\columnwidth,center}
    \begin{tabular}{ccccc}
    \toprule
    Methods & SR\% $\uparrow$ & CR\% $\downarrow$  & Time [ms] $\downarrow$ & MPL [rad] $\downarrow$ \\\midrule 
    NeHMO (ours) &  \textbf{92} & \textbf{2} & 36 ± 35 & 8.5 ± 4.6\\
    NeHMO (w/o Sym)    &  90 & 3 & 35 ± 33 & 8.3 ± 4.6 \\
    \Opt + DeepReach &  66 & 34 & 16 ± 4.1 &  6.2 ± 2.7 \\
        Ha et al. \cite{ha2020learningdecentralizedmultiarmmotion} & 79 & 13 & $\mathbf{<1.0}$ & \textbf{4.7 ± 1.3} \\ 
    CHOMP \cite{ratliff2009chomp} & 86 & 14 & 570 ± 270 & 6.7 ± 3.2 \\ 
    Naive Planner & 66 & 34 & $\mathbf{<1.0}$ & 6.2 ± 2.7\\
    \bottomrule
    \end{tabular}

\end{adjustbox}
    \caption{The success rate, collision rate, planning time, and mean path length of each method on the dual-UR5 system. Best results for each metric are shown in bold.}
    \label{tab:arm_planning_exp}
\end{table}

\begin{table}[b]
    \centering
    \begin{tabular}{ccccc}
    \toprule
    \# UR5 & Method & SR\% $\uparrow$ & CR\% $\downarrow$  & Time [ms] \\\midrule 
    \multirow{3}{*}{Three} & NeHMO & \textbf{86} & \textbf{5} & 41 $\pm$ 43 \\ 
    & CHOMP & 68 & 32 & 650 $\pm$ 300\\
    & Naive Planner & 59  & 41 & $\mathbf{<1.0}$ \\
    \midrule
    \multirow{3}{*}{Five} & NeHMO & \textbf{66} & \textbf{8} & 43 $\pm$ 52 \\ 
    & CHOMP & 32 & 68 & 1800 $\pm$ 1100 \\
    & Naive Planner & 38 & 62 & $\mathbf{<1.0}$ \\
    \bottomrule
    \end{tabular}

    \caption{The success rate, collision rate, and planning time of each method on multi-UR5 systems. Best results are shown in bold.
    }
    \label{tab:multi_ur5_exp}
\end{table}

This experiment also provides indirect validation of the learned value function for the dual-UR5 system. The \Opt +DeepReach baseline performs similarly to the naive planner, suggesting that the original DeepReach value network does not learn a sufficiently informative safety representation in this high-dimensional articulated setting. In contrast, incorporating the explicit boundary term substantially improves solution quality for NeHMO (w/o Sym).
Notably, the value function learned within NeHMO with symmetry reduction achieves performance comparable to NeHMO (w/o Sym), despite being trained with only half the data. This highlights the data-efficiency benefit of exploiting symmetry without sacrificing planning performance. An example of NeHMO successfully planning for the dual-UR5 system in both simulation and real-world experiments is shown in Figures~\ref{fig:planning_fig2} and~\ref{fig:intro_fig}.

\subsubsection{Multi-UR5 Scalability Study}
\begin{figure}[h]
    \centering
    \includegraphics[width=0.6\linewidth]{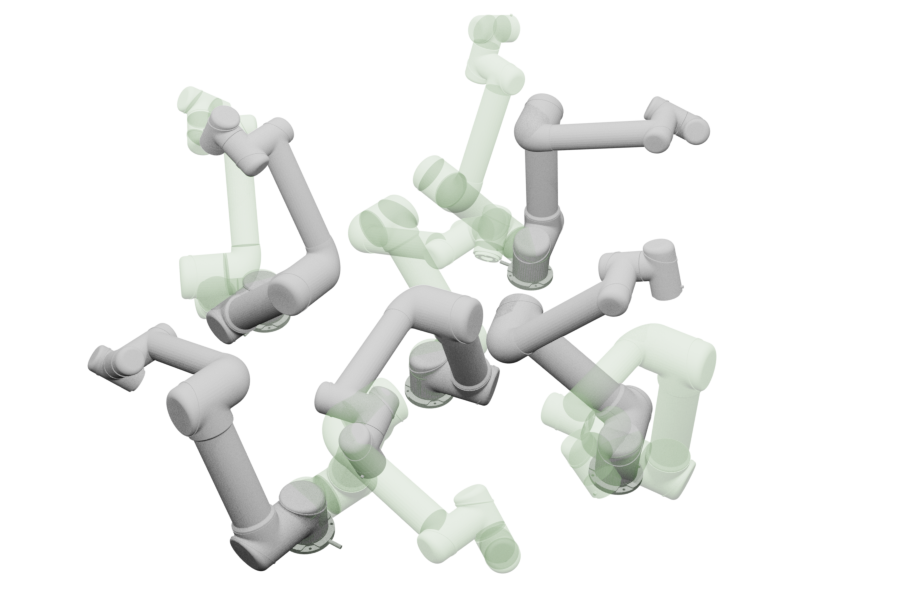}
    \caption{An example five-UR5 scenario.}
    \label{fig:five-ur5}
\end{figure}

 We further evaluate the scalability of NeHMO with an increasing number of manipulators in the workspace. To systematically increase task difficulty, a UR5 manipulator is placed in the center and surrounded by two or four additional UR5 manipulators. An illustration of a five-UR5 scenario is presented in Figure~\ref{fig:five-ur5}. Planning time is measured and averaged for this central manipulator only.

 Table~\ref{tab:multi_ur5_exp} reports the results. As the number of manipulators increases, the joint configuration space becomes more constrained due to the increase of inter-arm interactions, resulting in a high collision rate for both the naive planner and CHOMP. Despite this increased complexity, NeHMO maintains real-time performance and a low collision frequency. An example video of NeHMO planning for the five-UR5 setup is provided in the supplementary material.

 \subsection{UR5-Kinova Challenging Motion Planning Experiment}
Finally, to evaluate NeHMO in a heterogeneous setting, we consider a dual-arm system composed of a UR5 manipulator and a Kinova Gen3 manipulator. Similar to the dual-UR5 experiment, each arm is assigned randomly sampled start and goal configurations. To create a challenging coordination problem, configurations are selected such that simple linear interpolation between start and goal pairs would result in a collision. Successful execution therefore requires active coordination rather than independent goal-seeking behavior. 

We evaluate three planning scenarios:
\begin{itemize}
    \item \textbf{Naive / Naive}: Both manipulators move directly toward their goals without accounting for collisions. 
    \item \textbf{NeHMO / NeHMO}: Both arms use NeHMO for decentralized safety-aware planning. 
    \item \textbf{NeHMO / Naive}: The UR5 manipulator uses NeHMO while the Kinova arm follows a reckless policy that ignores collision avoidance. This setting evaluates NeHMO’s ability to generate defensive motion plans under unpredictable or unsafe agent behavior. 
\end{itemize}

Planning time is measured for manipulators that use NeHMO as the underlying planner. As shown in Table~\ref{tab:hetero_planning_exp}, the Naive / Naive baseline fails in all trials due to the challenging configurations selected. In contrast, when both arms employ NeHMO, the system achieves high success with a low collision rate. When the UR5 manipulator runs NeHMO while the Kinova arm executes a reckless policy, the UR5 is still able to avoid a substantial number of collisions. Under the selected start–goal configurations, the reckless policy can effectively behave as an adversary, as it may intrude into tightly shared workspace regions without coordination and may even hit the fixed base of the UR5 manipulator. Despite this, NeHMO demonstrates meaningful defensive capability under highly unpredictable conditions, significantly reducing collisions compared to the naive baseline.
An example of NeHMO successfully avoiding collision in the NeHMO / Naive experiment is presented in Figure~\ref{fig:ur5-kinova}. The accompanying video demonstration can be found in the supplementary material.

\begin{table}[t]
    \centering
    \begin{tabular}{ccccc}
    \toprule
    UR5 Planner & Kinova Planner & SR\% $\uparrow$ & CR\% $\downarrow$  & Time [ms] \\\midrule 
    Naive & Naive & 0  & 100 &  - \\
    NeHMO & NeHMO & 79 &   6 & 46 $\pm$ 46\\
    NeHMO & Naive & 47  & 39 & 41 $\pm$ 75 \\
    \bottomrule
    \end{tabular}
    \caption{Performance of each variant in the heterogeneous UR5–Kinova motion planning experiment.}
    \label{tab:hetero_planning_exp}
\end{table}

\begin{figure}[t!]
\centering
\hskip -1ex
\begin{subfigure}[t]{0.18\textwidth}
\includegraphics[width=1\textwidth]{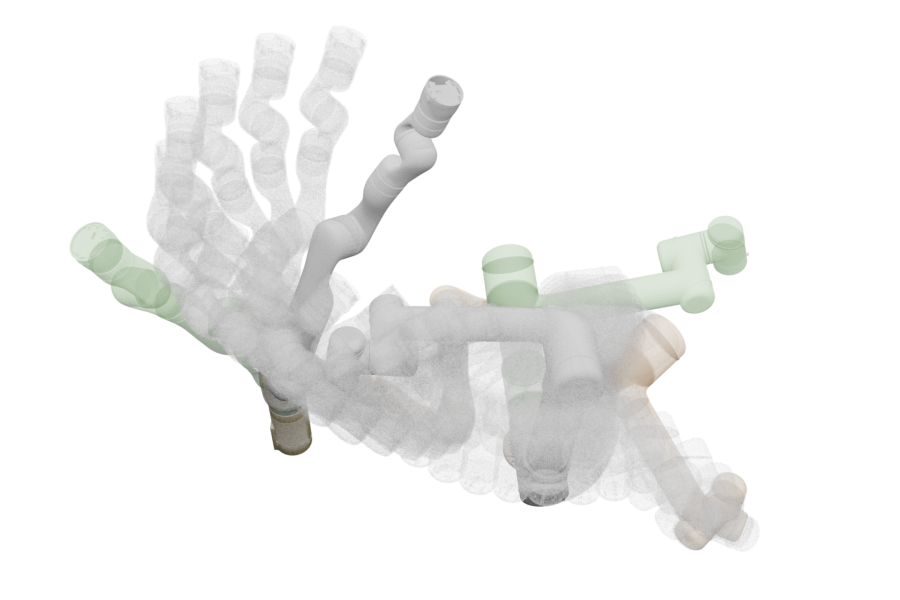}
\vskip -3ex
\caption{}
\end{subfigure}
\hskip -4ex
\begin{subfigure}[t]{0.18\textwidth}
\includegraphics[width=1\textwidth]{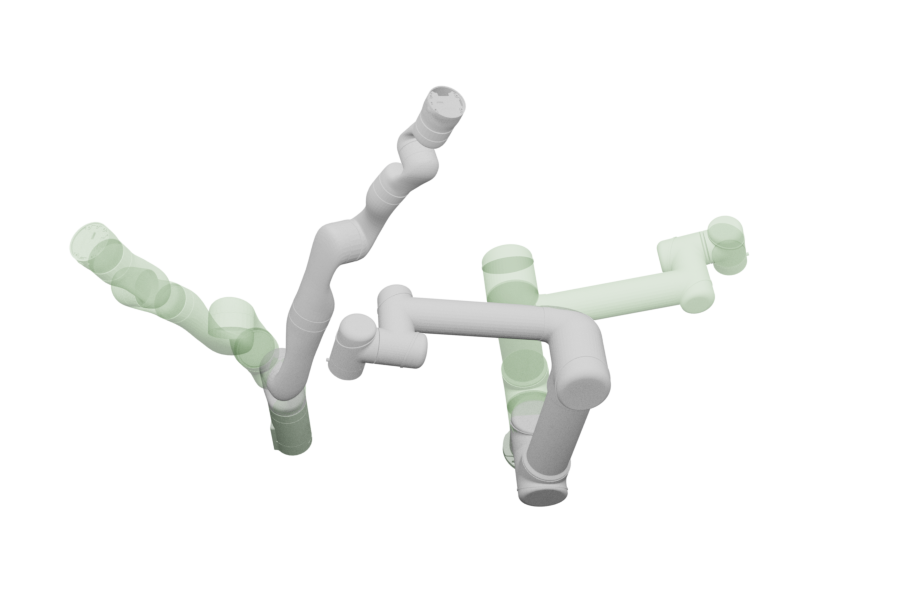}
\vskip -3ex
\caption{}
\end{subfigure}
\hskip -4ex
\begin{subfigure}[t]{0.18\textwidth}
\includegraphics[width=1\textwidth]{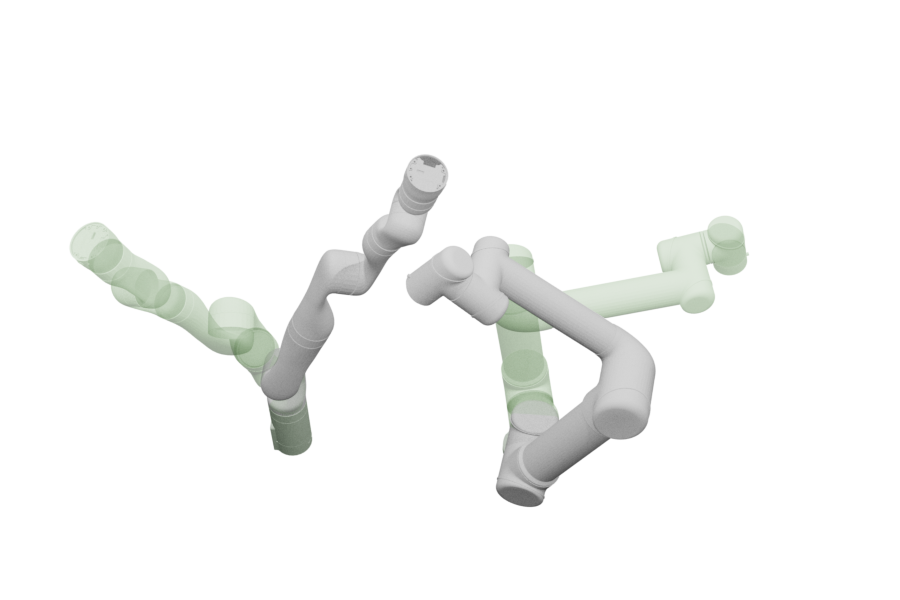}
\vskip -3ex
\caption{}
\end{subfigure}

\vskip -1ex
\hskip -1ex
\begin{subfigure}[t]{0.18\textwidth}
\includegraphics[ width=1\textwidth]{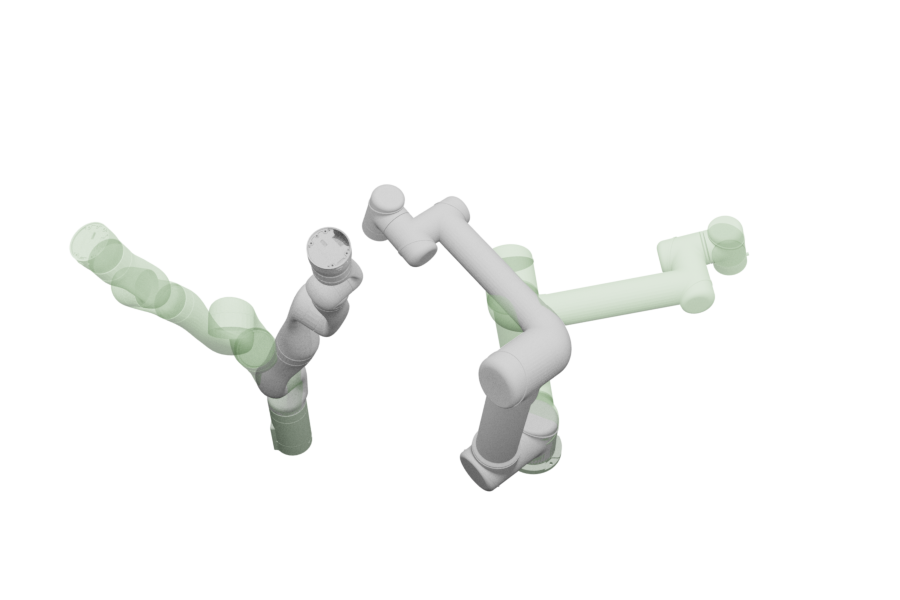}
\vskip -3ex
\caption{}
\end{subfigure}
\hskip -4ex
\begin{subfigure}[t]{0.18\textwidth}
\includegraphics[width=1\textwidth]{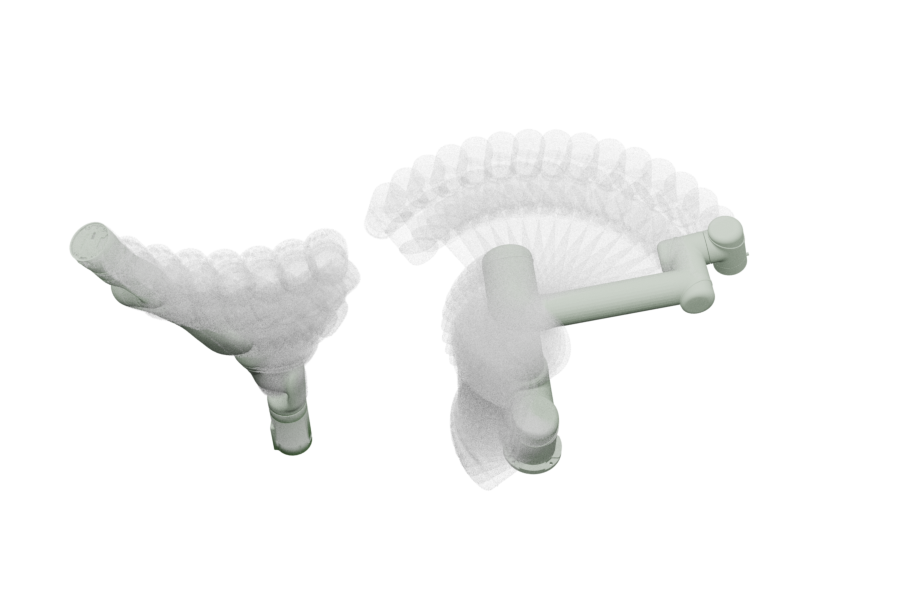}
\vskip -3ex
\caption{}
\end{subfigure}
\hskip -4ex
\begin{subfigure}[t]{0.18\textwidth}
\includegraphics[width=1\textwidth]{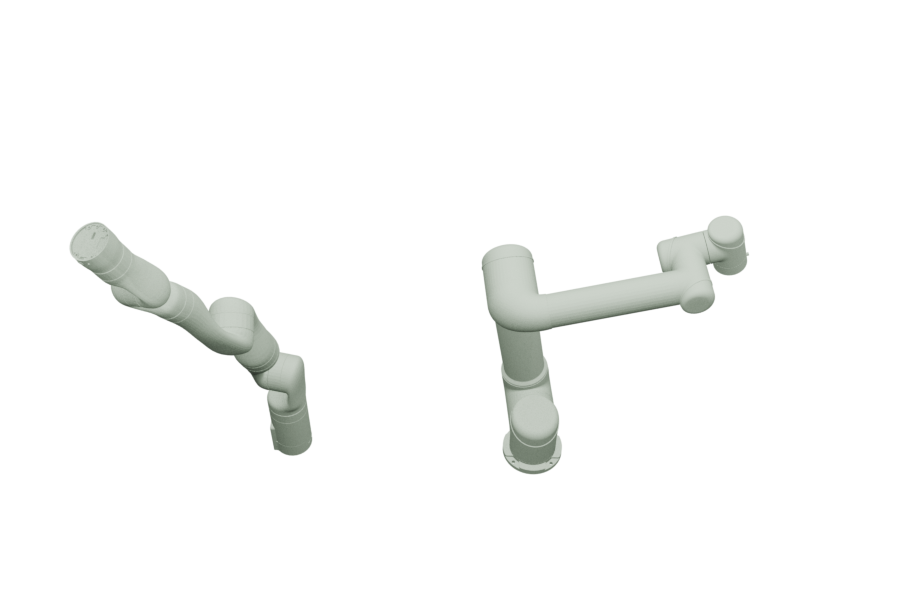}
\vskip -3ex
\caption{}
\end{subfigure}
\vskip -1ex
\caption{An example of NeHMO avoiding collision for the UR5 arm while the Kinova arm runs a {\color{red} reckless} policy in simulation. Current states are gray; intermediate trajectories are transparent; initial configurations are orange; goal configurations are green. The UR5 arm starts and gets close to the Kinova arm (a-b), adjust its configurations to avoid collision (c-d), and then safely reach its goal (e-f).}
\vskip -2ex
\label{fig:ur5-kinova}
\end{figure}

\section{Conclusion and Discussion} \label{sec:conclusion}

This manuscript proposes a framework to solve multi-arm motion planning problems with scalable HJR learning. We demonstrate its utility in solving HJR problems on multi-arm systems and present a trajectory optimization-based planner formulated with the learned HJR solution. We show that our framework solves challenging multi-arm motion planning tasks in real-time, is flexible with varying numbers of constraints, and efficiently scales with the number of robots in the system.

Future works aim to address several current limitations. The first is the loss of formal safety guarantees due to the use of neural networks. We see opportunities to amend the lost guarantees through post-training certification and conformal prediction for selecting the safety margin \cite{li2024certifiabledeeplearningreachability, lin2023generatingformalsafetyassurances}. Another limitation is the conservatism introduced in the HJI formulation and the worst-case prevention strategy, as empirically observed in the experiments from the increased path lengths and the trials that are neither a success nor a collision. This conservatism is governed by $t_{\text{plan}}$ and $t_{\text{safe}}$. Therefore, we aim to (i) accelerate solving \Opt to shorten the adversarial horizon (i.e., $t_{\text{plan}}$), and (ii) adopt a general-sum game formulation that accounts for other agents' intentions.  Finally, we plan to incorporate techniques such as prioritized planning and deadlock resolution strategies \cite{bakker2023multi} to extend our method to mobile manipulators, where the system can both be large-scale and have complex collision geometry.

\balance
\bibliographystyle{IEEEtran}
\bibliography{references}

\appendices
\section{Detailed System Setup} 
\label{sec:system-setups}

This section presents the details of the multi-robot systems studied for HJR modeling and their symmetry. Subscripts are used to index vector components. For example, the $j$-th component of the state of the $i$-th agent is denoted $x_j^i$. 

\subsection{Air3D}
The Air3D system models the interaction between two Dubins' cars and is a common benchmark for reachability analysis \cite{deepreach, mitchell2020robust}. It uses a relative dynamics defined by
\begin{align}
    \Dot{x}_{1} &= -v + v\cos x_{3} + \omega_e x_2\nonumber \\
    \Dot{x}_{2} &= v\sin x_{3} - \omega_e x_{1} \\
    \Dot{x}_{3} &= \omega_p - \omega_e, \nonumber
\end{align}
where $x_1$ and $x_2$ are the relative positions and $x_3$ is the relative heading. The superscripts to distinguish the robots are dropped in this instance of relative dynamics. The vehicles maintain a constant linear velocity $v$ and are controlled through $\omega_e, \omega_p \in [-\Bar{\omega}, \Bar{\omega}]\subset \mathbb{R}$ as their angular velocities.

The Air3D system exhibits symmetry in its dynamics and the boundary condition $l(x) = ||(x_1, x_2)||_2 - 2r$ where $r$ is the vehicles' radii. Consequently, the state mapping function $f(x_1, x_2, x_3) = (x_1, -x_2, -x_3)$ leads to the symmetry in the value function
\begin{align}
   & V(t, x_1, x_2, x_3) = V(t, x_1, -x_2, -x_3)\nonumber\\
   & \phantom{----}\forall t\in [0,T], x_1, x_2 \in \mathbb{R}, x_3 \in[-\pi,\pi].
\end{align}

\subsection{Simple Arm}
The simple arm system is composed of two 2-link planar manipulators in 2D space with fixed bases. We abstract the geometry of a manipulator to a chain of 2 links represented by line segments. The dynamics follow a velocity control scheme described by $q_j^i = u_j^i$ for $i=1,2,\ j=1,2$, where $u^i_j \in [-\Bar{u}, \Bar{u}]\subset \mathbb{R}$ is the joint velocity. The boundary condition function is based on the minimum pairwise distances between the links of the two manipulators. 

A symmetry in the value function is observed as
\begin{equation}
    V(t, x) = V(t, -x)\ \forall t\in [0,T], x\in[-\pi,\pi]^4.
\end{equation}

\subsection{Dual-Manipulator} \label{sec:ur5}

Similar to the simple arm system, the dynamics of the dual-UR5 manipulator system (i.e., the dual-UR5 system and the UR5-Kinova system) are $q^i_j = u^i_j\  \text{for}\ j=1,\dots,6,\ i=1,2$, where $u_j^i \in [-\Bar{u}, \Bar{u}] \subset \mathbb{R}$. The boundary condition defined for collision avoidance is a signed distance function between the manipulators, parameterized by their configurations. In practice, such a signed distance function and its gradient are hard to obtain due to the complex geometry of the manipulators. We therefore train a multi-layer perceptron (MLP) network to efficiently approximate this signed distance function similar to \cite{9976191}.

A symmetry in the dual-manipulator system is
\begin{align}
    & V(t, q^1_1, q^1_2, \dots, q^1_6, q^2_1, q^2_2, \dots, q^2_6) \nonumber\\ &= V(t, \pi - q^1_1, -q^1_2, \dots, -q^1_6, \pi-q^2_1, -q^2_2, \dots, -q^2_6)\nonumber \\  & \phantom{--} \forall t\in [0,T], q^i_{j}\in[-\pi,\pi], i=1,2,j=1,\dots,6.
\end{align}

\end{document}